# Goal-Driven Robotic Pushing Using Tactile and Proprioceptive Feedback

John Lloyd* and Nathan F. Lepora*

*Abstract*—In robots, nonprehensile manipulation operations such as pushing are a useful way of moving large, heavy or unwieldy objects, moving multiple objects at once, or reducing uncertainty in the location or pose of objects. In this study, we propose a reactive and adaptive method for robotic pushing that uses rich feedback from a high-resolution optical tactile sensor to control push movements instead of relying on analytical or data-driven models of push interactions. Specifically, we use goal-driven tactile exploration to actively search for stable pushing configurations that cause the object to maintain its pose relative to the pusher while incrementally moving the pusher and object towards the target. We evaluate our method by pushing objects across planar and curved surfaces. For planar surfaces, we show that the method is accurate and robust to variations in initial contact position/angle, object shape and start position; for curved surfaces, the performance is degraded slightly. An immediate consequence of our work is that it shows that explicit models of push interactions might be sufficient but are not necessary for this type of task. It also raises the interesting question of which aspects of the system should be modelled to achieve the best performance and generalization across a wide range of scenarios. Finally, it highlights the importance of testing on non-planar surfaces and in other more complex environments when developing new methods for robotic pushing.

## I. INTRODUCTION

Humans exploit both prehensile (e.g., grasping) and nonprehensile manipulation (e.g., pushing, pulling, rolling, pivoting) when working with objects. In robots, nonprehensile operations such as pushing are a useful way of moving large, heavy or unwieldy objects, moving many objects at the same time, or carrying out fine adjustments on small objects or assemblies [1], [2]. Pushing is also a useful way of reducing uncertainty in the location or pose of objects, particularly in cluttered environments [3], [4]. For some applications, the ability to push objects in a controlled way can help to reduce the size, complexity, power and cost of robots by removing the need for complicated and expensive hands or grippers.

However, robotic pushing presents several key challenges. Firstly, the physics of pushing is complicated to analyze because the distribution of support forces is mostly unknown and varies during the pushing operation. The analysis is made more difficult by the high degree of spatial, temporal and directional variability of the frictional forces [5]. Secondly, the control problem is hard to solve because it is nonlinear and subject to nonholonomic constraints (there are only specific directions that an object can be pushed in from a particular configuration). Finally, accurate information on the object state and motion can be difficult to extract from raw sensory data. Computer vision systems are good at providing global state information but can suffer from low accuracy, calibration errors, and occlusions [6]–[8]. Force/torque sensors and tactile sensors do not suffer from problems with occlusions but only provide local information on the object state.

In this study, we focus on the problem of how to use a robot arm to push an object from an arbitrary start position across a supporting surface to a final target position (but we do not address the related problem of how to orientate the object in a specific target pose). In contrast to most existing analytical and data-driven approaches, our method does not rely on explicit models of push interactions that predict how an object pose will change in response to a specific push action. Instead, it uses a reactive and adaptive procedure to explore around objects to find a stable pushing path. Specifically, we use goal-driven tactile exploration to actively search for stable pushing configurations that cause the object to maintain its pose relative to the pusher while incrementally moving the pusher and object towards the target. To facilitate this, we use feedback from an optical tactile sensor to predict the pose of the object relative to the pusher, and proprioceptive feedback from the robot to determine the pose of the pusher (end-effector) in the base frame. On flat surfaces, our method is robust to variations in contact position and angle, start position (relative to the target position) and different object shapes and sizes. The general nature of our method also enables us to push objects over curved surfaces, which to the best of our knowledge is the first time this has been done.

In this paper, we begin by reviewing existing methods for robotic pushing and, in particular, methods where tactile feedback has been used (Section II). We move on to describe our approach in detail, focusing on the robot/sensor system, the controller used for tactile exploration and target alignment, and the deep neural network component used to perceive the state of the pushed object (Section III). We then evaluate our method in three experiments (Section IV) and conclude by discussing our findings and their implications (Section V).

## II. BACKGROUND AND RELATED WORK

Most existing robotic pushing methods fall into one of two types. The first type relies on analytical, physics-based models of pusher-object interactions within higher-level planning and control frameworks. The second type uses data-driven methods to construct forward, or inverse models of pusher-object interactions, or to directly learn a pushing control policy.

Many physics-based, analytical models of robot pushing have been proposed over the past few decades. Mason [1] derived a simple rule, known as the *voting theorem*, for determining the direction of rotation of a pushed object. This rule only depends on the centre-of-mass of an object and not the underlying distribution of support forces. Building on results from classical plasticity theory, Goyal et al. [9] introduced the concept of the *limit surface* to provide a geometric description of how the motion of a sliding rigid body depends on the frictional load. Lee and Cutkosky [10] later went on to derive an *ellipsoid approximation* to the limit surface, to reduce the computational time needed to compute it. Lynch et al. [11] used the ellipsoidal approximation to obtain closed-form analytical solutions for sticking and sliding pushing interactions. Howe and Cutkosky [12] explored other geometric forms of limit surface and provided guidelines for choosing between them based on the pressure distribution, computation cost and required accuracy. Lynch and Mason [13] analyzed the mechanics, controllability and planning of pushing and developed a planner for finding *stable pushing* paths among obstacles.

* J. Lloyd and N. Lepora are with the Department of Engineering Mathematics, Faculty of Engineering, University of Bristol and Bristol Robotics Laboratory, Bristol, UK. e-mail {jl15313, n.lepora}@bristol.ac.uk.

Analytical approaches have the advantage of transparency and are based on relatively well-understood physics models whose parameters can be adjusted to suit the context. However, they can also cause difficulties if the underlying assumptions and approximations do not hold in practice. For example, many of them assume homogeneous, isotropic and stationary friction, which may not be valid for some surface materials [5].

In recent years, other researchers have turned to data-driven methods in an attempt to overcome some of the perceived problems with analytical models. Kopicki et al. [14] combine several regression models and density estimation models in a modular, data-driven approach for predicting the motion of pushed rigid objects. Hogan et al. [15] use Gaussian processes to model the change in position and orientation of an object in response to a push at a specified contact position and angle, and embed the model in a Model Predictive Control (MPC) framework. In an attempt to combine some of the strengths of analytical and data-driven methods, Zhou et al. [16] approximate the limit surface using a simple parameterized model (the level set of a convex polynomial) and fit the model using a computationally-efficient identification procedure. Other researchers have turned to deep learning to model the forward/inverse dynamics of pushed rigid body motion [17]–[20]; or to learn end-to-end control policies for pushing [21].

While data-driven approaches benefit from their flexibility and less restrictive assumptions, they can sometimes suffer from poor accuracy or data efficiency or may fail to generalize to new situations [2]. A comprehensive review and comparison of analytical and data-driven methods of robotic pushing can be found in [2].

### A. Use of Tactile Feedback in Robotic Pushing

Most contemporary methods for robotic pushing rely on feedback from computer vision systems or force/torque sensors. While computer vision systems are a useful way of obtaining *global* feedback for robotic tasks such as pick-and-place operations, they can pose challenges for tasks such as pushing, where sufficiently accurate and controlled interactions are needed. These challenges are compounded by calibration errors and occlusions [6]. For example, in their robotic pushing experiments, both Hermans [7] and Krivic [8] report computer vision problems due to noise, calibration errors, limited field of view and occlusions. Force/torque sensors are generally reliable and do not suffer from problems with occlusions, but during a single contact, they only provide limited *local* information on the object state [22]. Tactile sensors, particularly those that generate more informative, high-resolution output, may offer an alternative, or at least complement other approaches. However, to date, there has been relatively little research in this area.

Lynch et al. [11] were the first to demonstrate the use of tactile sensor feedback in robotic pushing. They used an optical waveguide tactile sensor mounted on one finger of a three-fingered hand to achieve stable translation of a rectangular object and circular disk on a moving conveyer belt. This operation can be viewed as equivalent to pushing an object in an allowed direction (subject to nonholonomic constraints and frictional forces at the point of contact) across a planar surface.

Jia and Erdmann [23] carried out a theoretical analysis of pushing with tactile feedback and showed that it is possible to determine the pose and motion of a planar object with known geometry solely from the tactile contact generated by pushing.

Another interesting approach that relies on tactile feedback [24] used an analytical model of planar sliding motion based on an ellipsoid limit surface to predict the outcome of push actions. Tactile feedback was provided by a set of torque sensors in the fingers of a Kuka/DLR robot. This method assumes that the object base shape is available as a model or can be estimated using a computer vision system. Exploratory robot movements are used to estimate the remaining model parameters: object weight, static and dynamic object-table and finger-object friction coefficients. They demonstrated their method by pushing a rectangular box across a flat uniform-friction tabletop. However, as discussed in the context of other analytical approaches above, it is not clear how well this method can generalize to new situations. In particular, they needed to make some fairly restrictive assumptions including uniform friction coefficients, point fingertip, uniform object-surface weight distribution, rectangular object-surface contact shape, and coincident centre-of-mass (CoM) and centre-of-friction (CoF).

More recently, Meier et al. [25] described a tactile-based method for pushing an object using frictional contact with its upper surface. Their method builds on earlier work where they used convolutional neural networks (CNN) to discriminate between translation, rotation and slip states in robotic manipulation tasks. They demonstrated their approach using a Shadow dexterous hand with high-speed, piezo-resistive tactile sensors to push an object slowly towards a wall until contact occurs. In comparison with this earlier work, our method uses a higher-resolution tactile sensor than has been used before to provide a rich feedback within a novel type of perception and control loop.

### B. Other Model-Free Feedback Control Approaches for Robotic Pushing

From a control perspective, the most similar approach to ours is that described by Hermans et al. [7]. In their work on autonomous learning of object affordances they describe two control strategies for pushing objects from a start position to a final target position on a flat horizontal surface. The first 'spin-correction' control strategy they describe is similar to the proportional-derivative (PD) style control law presented in [11]. As in our approach, their control strategy attempts to steer the object towards the target, while at the same time reducing the rotational motion of the object using a velocity error term. The second 'centroid alignment' control strategy counters the difficulty of estimating the orientation of rotationally symmetric objects by only using feedback on the relative location of the object's centroid and the assumed location of the contact point on the end effector.

The main difference between these two control strategies and ours is that they require the centroid or orientation of the object to be continually sensed and estimated using feedback from an RGB-D computer vision system (they describe several approaches for doing this using point cloud clustering, and fitting spherical or bounding box models). Their control approach also relies on *global properties* of the pushed objects (object centroid, object orientation), whereas as ours makes use of *local properties* that are sensed in an exploratory tactile servoing process as the pusher/robot traverses around the sides of the object. They demonstrated their approach using three types of motion primitives (overhead push, fingertip push, gripper sweep) on 15 everyday objects. Most of the objects were pushed to within 40 mm of the target, and some to within 20 mm using one of the two control strategies described above.

Another computer vision-based approach described by Krivic et al. [8] uses an adaptive feedforward/feedback controller to control the pushing moves of a mobile robot, while simultaneously adapting the control parameters in an online manner. Their system is more complex than ours and includes path planning and pushing strategy subsystems that compute collision-free 'pushing corridors' from a start point to finish point on a flat horizontal surface. Their control strategy is similar to the centroid alignment strategy used by Hermans et al. and, like theirs, it also assumes that the object centroid can be continually sensed and estimated using a computer vision system. However, one important difference is that their controller includes both an adaptive feedforward component (inverse model) and a feedback component, giving it the ability to gradually shift between slower initial feedback control, to faster feedforward control as the parameters of the inverse model are learned during early push interactions. The feedback component is essentially a proportional (P) controller, but with an added average-error term, which gives it similar characteristics to a proportional-integral (PI) controller. The adapted feedforward control parameters are the parameters of two 'activation functions' that determine the relative proportion of pushing and relocation movements that form the output of the inverse model. These parameters are adapted using a Bayesian estimation procedure.

They demonstrated their method using five geometric objects (two cylinders, two boxes and a ball) and seven fairly large irregular 'everyday' objects (five types of children's toy, a box and a grocery box). An object push sequence was considered successful if it approached within 50 mm of the target.

## III. METHODS

### A. Assumptions

We make two main assumptions in this work. Firstly, we assume *quasi-static interactions* where frictional forces dominate inertial forces. As well as simplifying the dynamics of the problem, this avoids having to deal with situations where the pusher loses contact with objects during push sequences. Secondly, we assume that we know or can measure the sensor contact angle in the vertical plane (or a plane that is normal to the supporting surface) needed to keep it moving parallel to the surface and avoid collisions with the surface or lose contact with the object. This second assumption is not required when pushing objects across a planar surface because we can restrict pusher movements to a fixed plane that is parallel to the surface by setting the corresponding gain terms to zero.

### B. Robotic System and Tactile Sensor

Our experimental system consists of a Universal Robotics UR5 6-DoF robotic arm, equipped with a TacTip optical tactile sensor, which we use to push objects across a surface (Figure 1a).

The TacTip sensor was developed by researchers at Bristol Robotics Laboratory (BRL) in 2009 [26] and has since been used in a wide variety of robotic touch applications [27]–[30]. Although we use the TacTip sensor in our work - primarily for historical reasons and its ready availability – a number of similar optical tactile sensors have been developed in recent years and we can see no reason why they could not be used with our method. These other optical-based tactile sensors are reviewed in [31], [32].

Our sensor tip has a gel-filled, hemispherical rubber skin (20 mm radius) with 127 internal markers arranged in a hexagonal grid pattern. While we have not investigated the extent to which the hemispherical geometry of the tip plays a role in this particular application, we suspect that sensors with flat pads would be less well-suited because of their more limited sensitivity to object orientation. An internal camera ($640 \times 480$ colour pixels at 30 fps) tracks the pin movements when the skin is deformed. The sensor can be mounted in two different

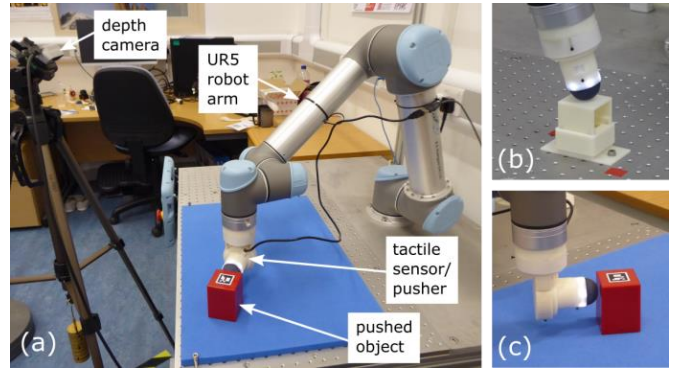

Figure 1. Robot system and tactile sensor/pusher. (a) General setup used for pushing experiments. (b) Sensor mounted in standard vertical configuration for collecting CNN training data. (c) Sensor mounted in right-angle configuration for robotic pushing.

configurations, depending on the operation being performed.

For collecting CNN training data (Section III.D), we attach the sensor in the standard vertical configuration (Figure 1b). For pushing, we mount it in a right-angle configuration to avoid problems with robot singularities (Figure 1c).

### C. Tactile Servoing and Target Alignment

We control pushing movements of the robot arm using a multiple feedback loop controller that contains two distinct control subsystems: *tactile servoing* and *target alignment* (Figure 2). These two subsystems work together to guide the pusher and object towards the target. The goal of the tactile servoing controller is to maintain the sensor/pusher at a fixed orientation and distance relative to the object during pushing moves. This first controller depends on feedback from a tactile perception subsystem that comprises an optical tactile sensor and a convolutional neural network (CNN). The CNN uses images from the tactile sensor to predict the sensor/pusher pose relative to the object. The goal of the second target alignment controller is to orientate the central axis of the sensor/pusher through the centre-of-friction towards the target by moving the sensor/pusher around the perimeter of the object during pushing moves. This second controller relies on proprioceptive feedback from the robot to determine the bearing and distance of the target relative to the sensor/pusher.

In describing the operation of the controller, we will use the following notation. We use capital letters to denote 3D coordinate frames. Specifically, we use the letter $B$ as the base frame of the robot arm, $S$ as the sensor/pusher frame, $F$ as the feature/contact frame, and $T$ as the target frame (whose axes are parallel to the corresponding base frame axes). During training, the feature/contact frame $F$ is fixed (relative to $B$) while $S$ is moved. During a push sequence, $F$ is determined (relative to $S$) by the outputs of the pose prediction CNN. The frames $S'$, $S''$ and $S'''$ denote the reference sensor frame, tactile servoing correction frame and target alignment correction frame, respectively (Figure 3). The poses of these frames are determined by the reference sensor pose and the output/correction signals produced by the two PID controllers.

We also use the uppercase, pre-superscript/post-subscript notation $^{A}T_{B}$ to denote an SE(3) $\subset \mathbb{R}^{4 \times 4}$ transform (i.e., a $4 \times 4$ homogeneous coordinate transformation matrix) that describes the geometric relationship between the two coordinate frames $A$ and $B$. Such a transform specifies how to convert a pose expressed in coordinate frame $B$ to a pose expressed in coordinate frame $A$. We also use the corresponding lowercase notation $^{A}t_{B}$ to denote an Euler vector parameterization of $^{A}T_{B}$, written $^{A}t_{B} = (x, y, z, \alpha, \beta, \gamma) \in \mathbb{R}^{6}$, where $(x, y, z)$ represents the translational component of the transform, and $(\alpha, \beta, \gamma)$ represents the rotational component, expressed in extrinsic-$xyz$ Euler notation. In Figure 2, we use 'e2t' symbols to indicate where the conversions from Euler parameterizations to the corresponding SE(3) transforms take place. Similarly, we use 't2e' symbols to indicate where conversions in the opposite direction take place. In our code, we use the Python `transforms3d` package to convert between the two representations. For all SE(3) transforms and their Euler parameterizations, we use letters that correspond to the role of the transform in a particular context. So, for example, we use $^{A}P_{B}$ to represent a pose, $^{A}E_{B}$ to represent an error and $^{A}U_{B}$ to represent a control signal (and the corresponding lower-case notation to represent their Euler parameterizations).

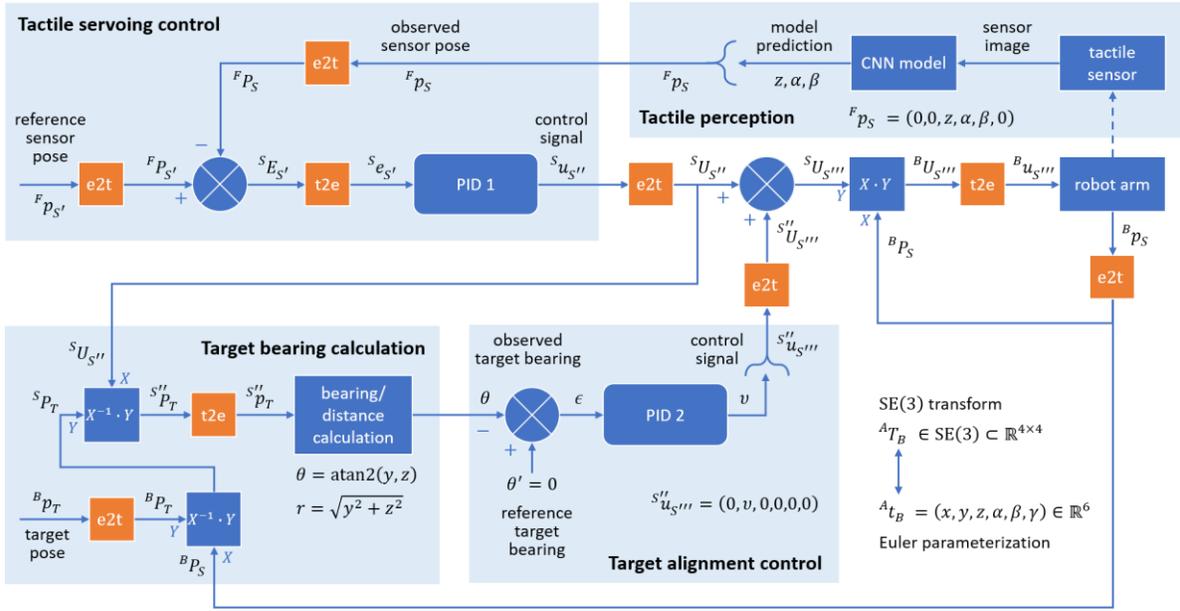

Figure 2. Multiple feedback loop controller used to control pushing movements. The first tactile servoing control loop maintains the pusher at a fixed orientation and distance from the sides of the object while pushing. The second target alignment control loop moves the pusher around the perimeter of the object to guide the pusher and object towards the target. Here, $S'$, $S''$ and $S'''$ denote sensor frames (see Figure 3 for more explanation).

The tactile servoing control subsystem shown in Figure 2 and described in this section is similar to the PID-based controllers we have used in previous work [29] and others have used in their work [33]. The goal of this controller is to maintain the sensor/pusher at a fixed orientation and distance relative to the pushed object. Equivalently, the controller tries to drive the error $^{S}E_{S'}$ between the observed sensor pose $^{F}P_{S}$ and a reference sensor pose $^{F}P_{S'}$ to zero (both poses being specified with respect to a common feature/contact frame). A key difference to earlier approaches is that here we compute the error using full SE(3) poses, rather than by subtracting their corresponding vector parametrizations. So, the error is computed as $^{S}E_{S'} = {}^{S}P_{F} \cdot {}^{F}P_{S'} = {}^{F}P_{S}{}^{-1} \cdot {}^{F}P_{S'}$, where the product and inverse operations are implemented using standard matrix products and inverses. The SE(3) error $^{S}E_{S'}$ is then converted back to its Euler parametrization $^{S}e_{S'}$ before being passed to a (6-channel) vector PID controller. The controller tries to reduce the error to zero by computing a correction $^{S}U_{S''}$ to the sensor pose using the following PID control law:

$$^{S}u_{S''}(t) = K_{p}\,{}^{S}e_{S'}(t) + K_{i}\int_{0}^{t}{}^{S}e_{S'}(t')dt' + K_{d}\,{}^{S}\dot{e}_{S'}(t). \quad (1)$$

In this equation, the proportional, integral and derivative error vectors are multiplied by the respective gain coefficient matrices $K_{p}, K_{i}, K_{d} \in \mathbb{R}^{6 \times 6}$, which were hand-tuned in a trial-and-error procedure (Table 1). The integrated error is also clipped to avoid 'integrator windup' effects. By treating the controller error as a regular 6-element vector within the PID control law (Equation 1) we avoid the complication of integration and differentiation over SE(3). While there has been some research to address this problem [34], [35] we view this as a higher-order optimization that is unlikely to have a significant effect on performance for this application.

The $z$, $\alpha$, and $\beta$ components of the reference sensor pose, $^{F}p_{S'}$, were set as follows. The $z$-component that determines the sensor contact depth was set to $z = 2$ mm. The $\alpha$-component that determines the sensor contact angle in the horizontal plane (or a plane that is tangential to the supporting surface) was set to $\alpha = 0$ to ensure that the sensor remains perpendicular to the pushed edge. The $\beta$-component that determines the sensor contact angle in the vertical plane (or a plane that is normal to the supporting surface) was set to $\beta = 0$ for the objects used in this study. The remaining $x$, $y$ and $\gamma$ components were set to zero as they are not required for this particular task.

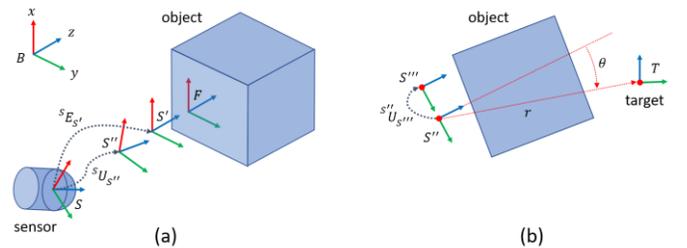

Figure 3. Target bearing calculation in the tactile servoing correction frame. (a) The sensor pose correction $^{S}U_{S''}$ produced by the tactile servoing control loop implicitly defines an intermediate sensor frame $S''$ (the tactile servoing correction frame), lying somewhere between the current sensor frame $S$ and the target frame $S'$. (b) The target bearing $\theta$ is computed in the $yz$-plane of the tactile servoing correction frame $S''$.

| PID | Parameter | Value |
|---|---|---|
| 1 | $K_p$ | diag$(0, 0, 0.9, 0.9, 0.9, 0)$ |
| | $K_i$ | diag$(0, 0, 0.1, 0.1, 0.1, 0)$ |
| | $K_d$ | $\mathbf{0}$ |
| | $\int$ error clip ranges | $[-5, 5]$ (translation) |
| | | $[-25, 25]$ (rotation) |
| 2 | $k_p$ | 0.2 |
| | $k_i$ | 0 |
| | $k_d$ | 0.5 |
| | output clip range | $[-5, 5]$ |

The tactile servoing controller output ${}^S u_{S''}$ is then converted back to an SE(3) transform ${}^S U_{S''}$ and used, together with the sensor pose ${}^B P_S$ provided by the robot arm, to calculate the bearing $\theta$ and distance $r$ of the target in the tactile servoing correction frame $S''$ (Figure 2). More specifically, the target pose, ${}^B P_T$ is first transformed to the tactile servoing correction frame $S''$ defined by the correction ${}^S U_{S''}$: ${}^{S''} P_T = {}^S U_{S''}^{-1} \cdot {}^B P_S^{-1} \cdot {}^B P_T$. The transformed target pose, ${}^{S''} P_T$, is then converted to its Euler parameterization ${}^{S''} p_T$ and used to calculate the target bearing $\theta$ in the tactile servoing correction frame using the atan2 function (Figure 3): $\theta = \mathrm{atan2}\left({}^{S''} p_T^{(y)}, {}^{S''} p_T^{(x)}\right)$. Here, the $(y)$ and $(z)$ superscripts on the Euler parameterization indicate the $y$ and $z$ components of the parameterization, respectively. The target bearing is then passed to the tactile alignment control subsystem for further processing.

Unlike the tactile servoing subsystem, the target alignment control subsystem is based on a more conventional, scalar-valued PID controller. The goal of this controller is to orientate the central axis of the sensor/pusher (i.e., the pushing direction) through the centre-of-friction (CoF) towards the target by moving the sensor/pusher around the perimeter of the object during pushing moves. Equivalently, it tries to drive the error $\epsilon = \theta' - \theta$ between the observed target bearing $\theta$ and the reference target bearing $\theta'$ to zero, while simultaneously doing the same to $\dot{\epsilon}$ (this error derivative term will only be zero when pushing is directed through the centre-of-friction).

The target alignment controller tries to achieve this goal by producing an additional correction, ${}^{S''} u_{S'''} = (0, v, 0, 0, 0, 0)$, to the original, tactile servoing correction, ${}^S u_{S''}$. The $y$-component, $v$, of this correction is given by the PID control law:

$$v(t) = k_p \epsilon(t) + k_i \int_0^t \epsilon(t') dt' + k_d \dot{\epsilon}(t). \quad (2)$$

The proportional, integral and derivative scalar error terms are multiplied by the respective gain coefficients $k_p$, $k_i$ and $k_d$, which were hand-tuned in a trial-and-error procedure (Table 1). The controller output is also clipped to avoid losing contact with the object in situations where the bearing error is very large.

A key feature of the target alignment controller is that a non-zero target bearing error, $\epsilon$ produces a correction that moves the sensor/pusher around the perimeter of the object to reduce the bearing error on the next push movement (Figure 4.a). At the same time, a non-zero target bearing error derivative $\dot{\epsilon}$ produces a correction that encourages the sensor/pusher to push through the object centre-of-friction, which in turn reduces the amount of object rotation on the next push movement. This produces a similar effect to the PD- style control law used in [11] and the 'spin-correction' control strategy used in [7].

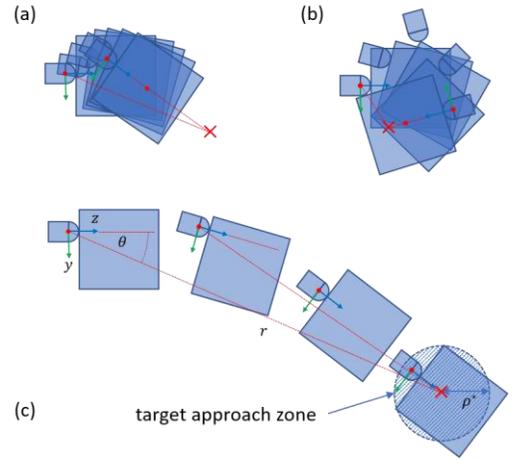

Figure 4. Target alignment control principles. (a) Reducing the target bearing error to zero under normal operation. (b) Larger perimeter moves required as object CoF moves past target. (c) Disengagement of target alignment control in target approach zone (radius $\rho^*$). Note that the sensor coordinate frame (shown) is located at the center of the hemispherical tip.

The target alignment controller output/correction is then converted to an SE(3) transform ${}^{S''} U_{S'''}$ and combined with the output/correction of the tactile servoing controller ${}^S U_{S''}$ to produce a composite correction ${}^S U_{S'''} = {}^S U_{S''} \cdot {}^{S''} U_{S'''}$. The composite correction is transformed from the sensor frame to the robot base frame using the end-effector pose, ${}^B P_S$ obtained from the robot: ${}^B U_{S'''} = {}^B P_S \cdot {}^S U_{S'''}$. The 'robot' in the figure is an abstraction of the physical robot and low-level control software in the sense that when provided with a base frame pose, it moves the end effector to the specified pose and makes a tapping movement along the central axis of the sensor, 10 mm forward and 5 mm backward.

At the end of a push sequence, as the object centre-of-friction (CoF) approaches the target and moves past it, larger moves around the object perimeter may be needed (Figure 4.b). These larger moves can result in push stability or controllability issues, or the sensor losing contact with the object. To get around this problem, we disengage the target alignment control inside a *target approach zone* of radius $\rho^* = 60$ mm (Figure 4.c). In our experiments, the radius of the target approach zone was chosen to be large enough that the CoF of the biggest object could not be moved past the target for any stable pushing configuration. While this is a reasonably crude solution, we found that performance is adequate for an initial study, and it allows us to focus on our main contribution of pushing control. The push sequence is terminated when the target is closer than 20 mm from the centre of the sensor tip (radius = 20 mm). This condition ensures that the sensor contact point is as close as possible to the target on termination (given the final orientation of the sensor and the length of the push steps), without moving too far past it.

### D. Tactile Pose Estimation

As in previous work [29], [30], we use a convolutional neural network (CNN) to predict the relative distance and orientation of the optical tactile sensor with respect to the (local) object surface, based on an input image received from the sensor (Figure 5). While we use the TacTip sensor in our work, in principle, any optical tactile sensor that provides enough information to predict the state variables could be used here, e.g., GelSight [36]. Alternatively, if a non-optical tactile sensor were used, the CNN could be replaced with a neural network or another nonlinear regression model that is capable of learning the

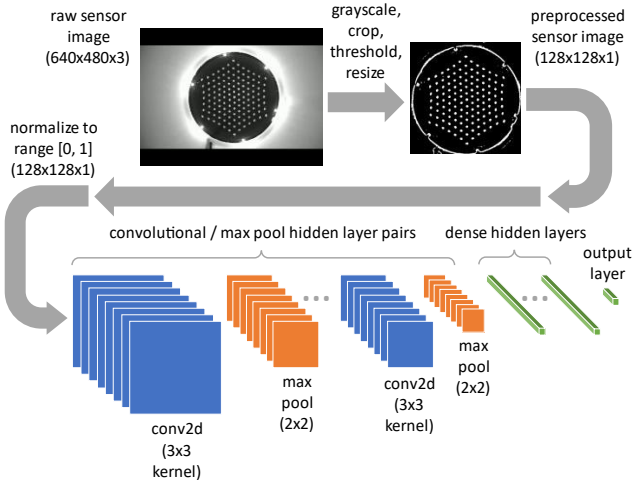

Figure 5. Convolutional neural network (CNN) and preprocessing used to predict sensor pose relative to the object surface.

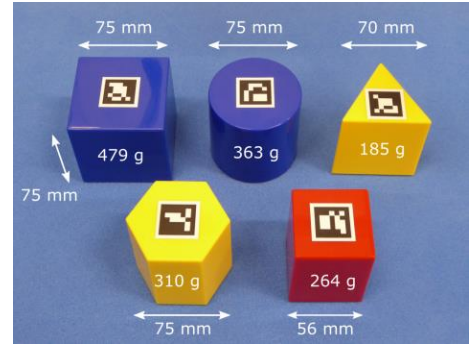

Figure 6. Prism-shaped plastic objects used in Experiments 1 and 2.

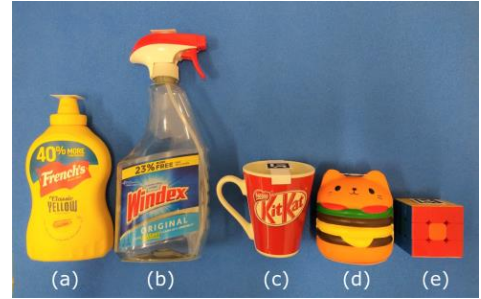

Figure 7. Irregular everyday objects used in Experiment 3: (a) plastic mustard bottle, (b) transparent spray bottle, (c) ceramic mug, (d) foam soft toy, and (e) Rubik's cube

mapping between sensor outputs and pose state variables.

Using our CNN-based approach, the $640 \times 480 \times 3$ tactile sensor image is first converted to grayscale, then cropped, thresholded and resized to $128 \times 128 \times 1$ before being normalized to floating-point numbers in the range $[0,1]$. We use thresholding to reduce the sensitivity of the model to variations in ambient lighting and contact surface patterning. The preprocessed image is then applied as input to a CNN that predicts the contact angles and depth.

The CNN has a standard 'LeNet-style' architecture and is built from a stack of convolutional layers ($3 \times 3$ kernels, stride 1, 'same' padding) interspersed with max-pooling layers ($2 \times 2$, stride 2). The output of the convolutional base feeds into a densely-connected regression network with three outputs. The three outputs form the $z$, $\alpha$, and $\beta$ components of the sensor pose, $^{F}P_{S}$, relative to the point of contact with the object (assuming extrinsic-$xyz$ Euler convention). The $x$, $y$, and $\gamma$ components of the sensor pose are all set to zero. The CNN hyperparameters were optimized with the Bayesian Optimization and Hyperband (BOHB) optimizer [37], using procedures that are largely identical to those used in our previous work - see [29], where they are discussed in depth. As CNN hyperparameter optimization is not the main focus of this work, we just report the values of the tuned parameters (Table 2).

Two data sets were used for training and validating the model, and a third data set held back for final testing before deployment. Each data set contained 2,000 labelled sensor images, which we collected by bringing the sensor into contact with a flat training surface at quasi-random poses (Sobol sampling) in the ranges $z \in [1,5]$ mm, and $\alpha, \beta \in [-20, 20]$ degrees. During data collection, rather than move the sensor into position from above, we moved it into place by sliding/rolling it across the surface from a randomly sampled offset. We found that this improves generalization performance and avoids the need for complex data augmentation to simulate the effects of motion-dependent shear. We trained the model from a random weight initialization using the Adam optimizer (learning rate $= 10^{-4}$; decay $= 10^{-6}$) for 100 epochs or until the mean-squared error (MSE) validation loss showed no improvement for ten consecutive epochs. After training, the model delivered test predictions for $z$, $\alpha$ and $\beta$ with mean absolute errors (MAE) 0.1 mm, $0.39°$ and $0.34°$, respectively. This is similar to the accuracy reported before in [29].

TABLE 2
Optimized CNN hyperparameter values

| CNN hyperparameter | Optimized value |
|---|---|
| # convolutional/max-pool hidden layer pairs | 7 |
| # filters per convolutional layer (all layers) | 248 |
| # dense hidden layers | 3 |
| # hidden units per dense layer | 244 |
| activation function (all hidden layers) | ELU |
| batch normalization (convolutional layers) | None |
| dropout coefficient (dense layers) | 0.046 |
| L1 regularizer coefficient (dense weights) | $1.6 \times 10^{-6}$ |
| L2 regularizer coefficient. (dense weights) | $2.2 \times 10^{-6}$ |
| batch size | 16 |

It is important to note that the CNN was only trained once, and the model was then used for all objects and surfaces considered here: it did not need to be retrained or replaced for different-shaped objects with different friction properties (as we see later with a variety of irregular objects).

### E. Objects and Surfaces

In the first two experiments, we used a selection of prism-shaped plastic objects of different shape, size and mass (Figure 6). By using prism-shaped objects, we can simplify the calculation of the sensor contact angle, $\beta$, needed to keep push movements parallel to the supporting surface. This angle is typically set to zero since the sides of the objects are perpendicular to the supporting surface. However, it can also be varied to compensate for any forward tilt of objects on softer surfaces. For objects with sloping sides, such as cones or frustums, the sensor contact angle would need to be obtained from a model or measurements, or estimated by placing the tactile sensor in contact with the surface and the sides of the object.

In the third experiment, we used a selection of irregular everyday objects to demonstrate the robustness of the method to variations in object, shape, size and material (Figure 7). These objects were selected to illustrate a wide range of material properties such as shape, texture, compliance and transparency, using a manageable number of objects. Since our current design of tactile sensor (Figure 1) cannot be positioned close to the surface, we did not consider low-profile, flat objects. Furthermore, only the flat horizontal surface was considered here because, for irregular objects, curved surfaces present the additional challenge of pitching and colliding with the surface (this was prevented here by zeroing the corresponding elements of the tactile servoing gains, i.e., the fifth diagonal elements of $K_p$ and $K_i$ in Table 1). Overall, this was sufficient to explore the pushing control for irregular objects. Future work can build on this approach using a redesigned sensor to demonstrate more generality.

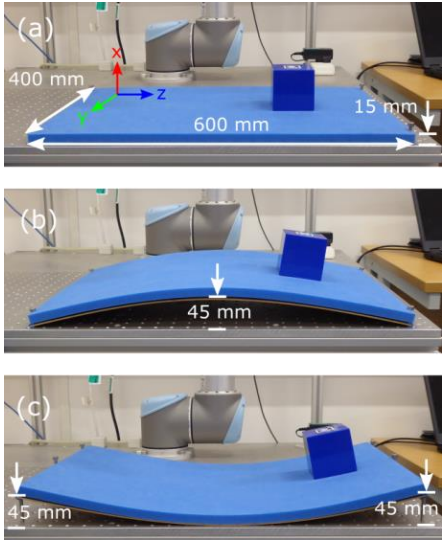

Figure 8. Polyethylene foam surfaces used in pushing experiments: (a) horizontal planar surface, (b) convex curved surface, and (c) concave curved surface. The position of the work coordinate frame used in all experiments is shown in (a).

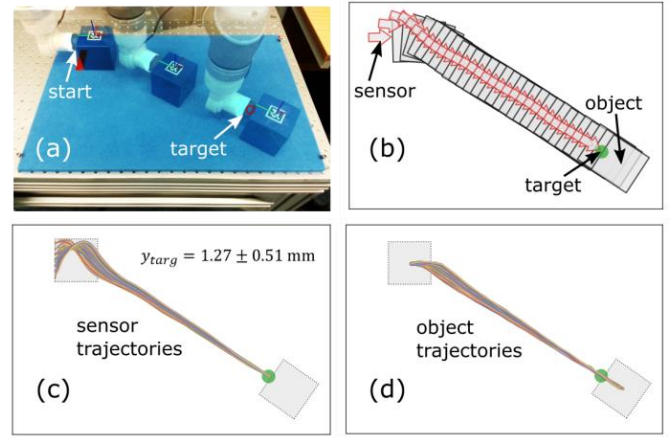

Figure 9. (a) Timelapse images and (b) sensor/object trajectories obtained when pushing the blue square prism from start position at work frame origin, over a horizontal planar surface towards target (green circle). (c) Converging sensor trajectories and (d) object trajectories for different initial contact positions and angles. The sensor is initialised from 7 different contact positions and 3 different angles, and 10 trials are performed for each of the 21 combinations. Also shown in (c) is the mean $\pm$ standard deviation of the distance $y_{targ}$ from the central axis of the sensor to the target on completion, for all 210 trials.

We push the objects across a medium-density, closed-cell, polyethylene foam sheet which can be laid flat as a horizontal plane or bent over a hardboard former to create a convex or concave curved surface (Figure 8). We selected this material because of its wide availability, its reconfigurability, and the fact that its softer nature make it one of the more challenging types of surfaces for pushing objects across [5]. For convenience, we also define a work coordinate frame (see figure) located at $(-85, -330, 70, 180, -90, 0)$ in the base frame. Here, the first three spatial coordinates are in mm; the last three rotation angles (extrinsic-$xyz$ Euler convention) are in degrees.

### F. Object Pose Tracking

To capture the motion of objects during push sequences, we use an Intel RealSense D435 depth camera to track ArUco markers ($25 \times 25$ mm), which are attached to the upper surface of each object. The camera is calibrated using intrinsic parameters obtained directly from the camera API, and extrinsic parameters calculated using a hand-eye calibration procedure. We use the colour images from the camera to calculate the (approximate) poses of markers in the camera frame, and depth readings taken at the marker centroids to compute their positions to greater accuracy. The marker poses and positions are transformed to the robot base frame using the extrinsic camera parameters, to give a positional accuracy of ~1 mm.

## IV. RESULTS

We evaluate our method in three experiments: the first investigates how robust the method is to variations in initial contact position and angle; the second evaluates how the method performs for five different object shapes and three different start positions (relative to the target) on planar and curved surfaces; and the third evaluates the robustness of the method for five irregular everyday objects by pushing them across a planar surface from the second start position (which we believe to be the most challenging of the three start positions).

### A. Experiment 1 – Varying Initial Contact Position and Angle for a Horizontal Planar Surface

In the first experiment, we pushed the blue square prism from the start position at the origin of the work coordinate frame to the target position at $(0, 200, 400, 0, 0, 0)$ (Figure 9a and Figure 9b). To investigate how the method responded to variations in initial contact position and angle we initiated the push sequence from seven different spatial offsets and three different angular offsets along the side of the object closest to the sensor. We used spatial offsets in the range $(-30, -20, -10, 0, 10, 20, 30)$ mm and angular offsets in the range $(-20, 0, 20)$ degrees and recorded the sensor and object (ArUco marker) trajectories for ten trials of each combination of spatial and angular offset (Figure 9c and Figure 9d). We also calculated the mean $\pm$ standard deviation of the distance $y_{targ}$ from the central axis of the sensor to the target on completion of the push sequence as a measure of how close the pusher was able to approach the target. In this experiment, we calculated these values to be $1.27 \pm 0.51$ mm over the $7 \times 3 \times 10 = 210$ trials.

Despite variations in the initial contact position and angle, the trajectories all converge smoothly on the target with good accuracy, confirming that the method is robust to variations in the initial configuration.

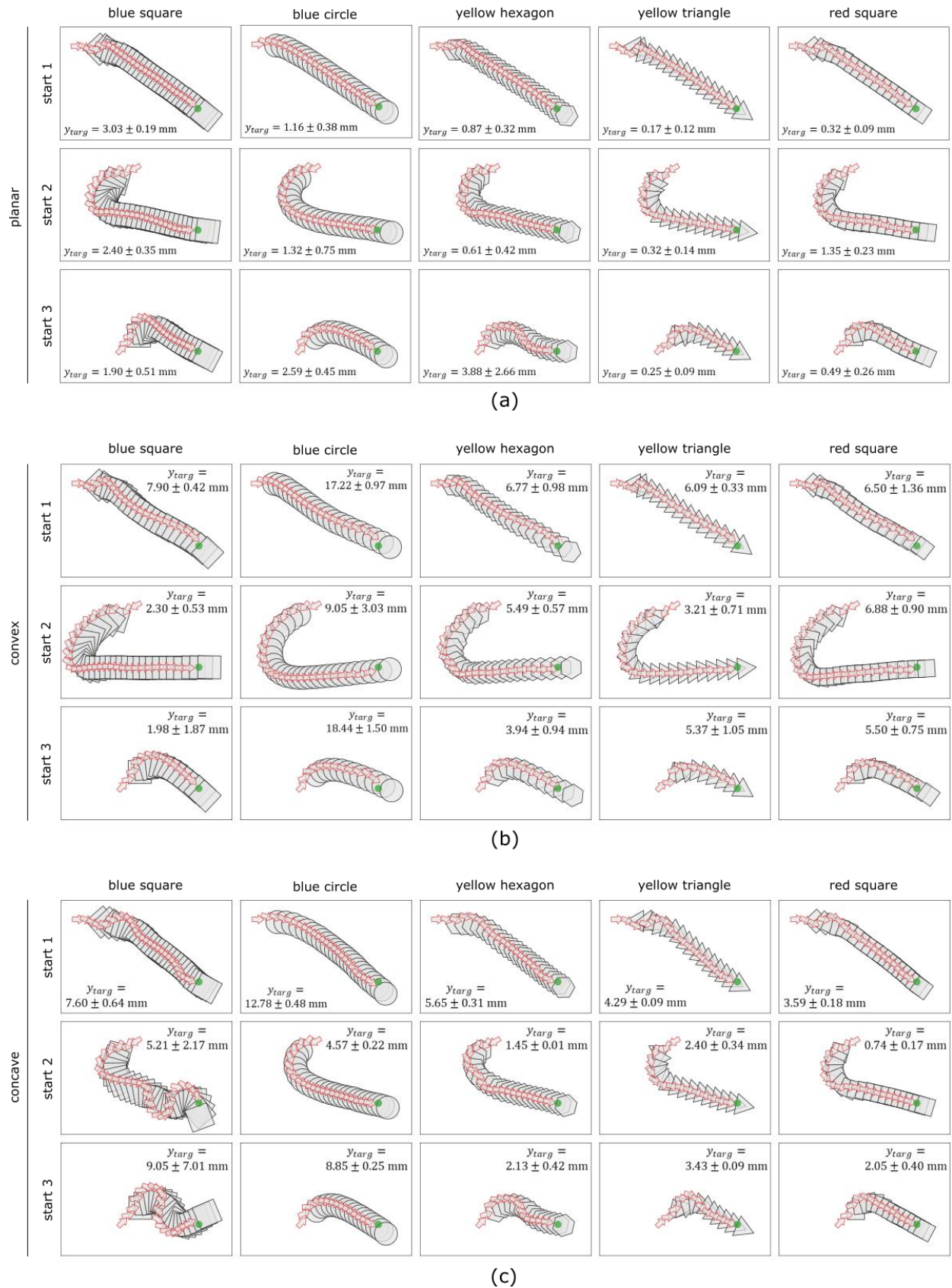

Figure 10. Examples of sensor and object trajectories obtained when pushing five geometric objects from three different start positions over three surfaces towards the target (green circle): (a) horizontal planar surface, (b) convex surface, and (c) concave surface. Also shown is the mean ± standard deviation of the distance $y_{targ}$ from the central axis of the sensor to the target on completion, taken across all corresponding trials.

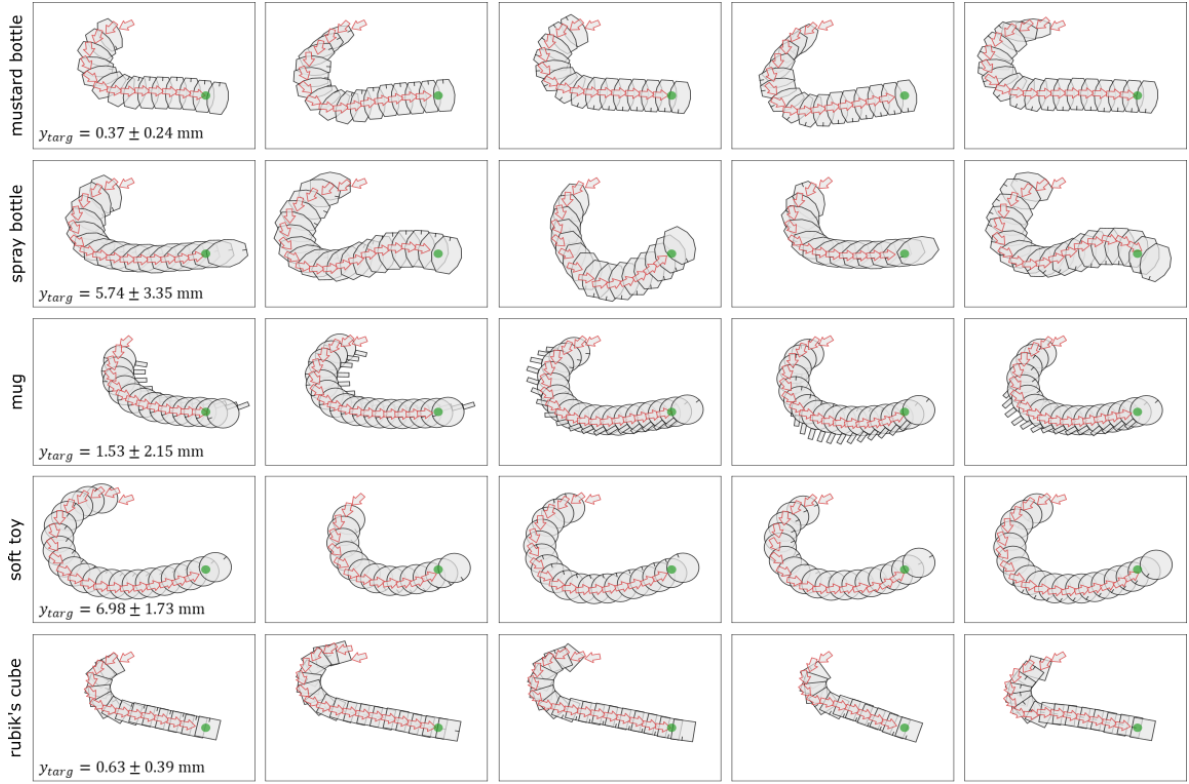

Figure 11. Examples of sensor and object trajectories obtained when pushing five irregular everyday objects from five random initial orientations at the second start position ('start 2') towards the target (green circle). Also shown in the first column is the mean ± standard deviation of the distance $y_{targ}$ from the central axis of the sensor to the target on completion, taken across all 10 corresponding trials. The position and orientation of objects during push sequences is indicated by the outline (viewed from above) and the short indicator line/notch bordering the edge of the objects.

### B. Experiment 2 - Varying Object Shape and Start Position Relative to Target for Planar and Curved Surfaces

In the second experiment, we pushed five geometric objects from three different start positions towards the target at $(0, 200, 400, 0, 0, 0)$ (Figure 10). For the first start position (labelled 'start 1' in the figure), the sensor was initially moved to the origin of the work frame, as it was for the first experiment. For the second start position ('start 2'), the sensor was initialized at pose $(0, 0, 200, -150, 0, 0)$, which is closer to the target but pointing in the opposite direction so that *more extensive* manoeuvring is required in order to reach it. For the third start position ('start 3'), the sensor was initialized at pose $(0, 200, 150, 45, 0, 0)$, which is even closer to the target but pointing away from it so that *more rapid* manoeuvring is required in order to reach it. For each start position, the objects were placed directly in front of the sensor tip and rotated into the least favourable (unstable) configuration such that an external corner of the object was centred on the sensor tip. For the curved surfaces, we increased the height of the start pose to allow for the additional 45 mm surface height at the centre of the convex surface and at the edges of the concave surface.

For each of the $5 \times 3 = 15$ combinations of object and start position, we recorded the sensor and object trajectories over ten trials and calculated the mean ± standard deviation of the distance $y_{targ}$ as a measure of how close the pusher was able to approach the target. These values are displayed alongside the corresponding sensor and object trajectories in Figure 10. The trajectories are plotted for every five push movements to avoid the figures becoming too cluttered. Across all objects and start positions, the mean ± standard deviation of the distance $y_{targ}$ is $1.18 \pm 1.10$ mm. For the convex and concave curved surfaces, these values increase to $5.54 \pm 4.17$ mm and $4.78 \pm 3.35$ mm, respectively.

### C. Experiment 3 – Pushing Irregular Everyday Objects on Planar Surfaces

In the third experiment, we pushed five irregular everyday objects across a flat surface from random initial orientations at the second start position ('start 2') towards the target at $(0, 200, 400, 0, 0, 0)$ (Figure 11). Since, in this experiment, we also varied the initial orientations of objects, we decided to limit the number of start positions to a single one to avoid too many experimental configurations. In any case, we feel that the second start position is the most challenging one and is sufficient to demonstrate that the method can cope with pushing a range of irregularly-shaped everyday objects. Our choice of irregular objects and other experimental conditions are explained in Section III.E of the methods.

For each object, we recorded the sensor and object trajectories over ten trials (each starting from a different random object orientation) and calculated the mean ± standard deviation of the distance $y_{targ}$ as a measure of how close the pusher was able to approach the target. These values are displayed alongside the corresponding sensor and object trajectories in the first column of Figure 11. Across all objects and (random) start orientations, the mean ± standard deviation of the distance $y_{targ}$ is $3.05 \pm 3.36$ mm.

## V. Discussion

In this paper, we have proposed a method of robotic pushing based on goal-based tactile exploration. We have shown that it can achieve good accuracy and is robust to variations in initial contact position and angle, and to different object shapes and start positions relative to the target. When pushing regular geometric objects across a flat, horizontal surface, the mean $\pm$ standard deviation of the distance $y_{targ}$ from the central axis of the sensor to the target on completion of the push sequences is $1.18 \pm 1.10$ mm. The method, therefore, works highly accurately.

The method also generalizes to curved surfaces but has lower performance. For the convex surface, the mean $\pm$ standard deviation of $y_{targ}$ increases to $5.54 \pm 4.17$ mm; for the concave surface, it increases to $4.78 \pm 3.35$ mm. The deterioration in performance observed for the curved surfaces is probably due to several factors. Firstly, on sloping surfaces, it may not be possible to satisfy all of the controller goals simultaneously and hence some degree of tradeoff may be needed. This might explain the much weaker performance of the circular prism when compared to the other shapes on the curved surfaces. If the centre-of-friction (CoF) does not lie on the central axis, there is (in general) no point on the perimeter where the target bearing is zero and where the sensor is perpendicular to the sides and pushing through the CoF. Secondly, the surface curvature can introduce second-order effects, which makes it more difficult for the controller to correct the error: the first-order error derivative is already used to discourage the object from rotating during push movements on flat or inclined planes, so a second-order error derivative may be needed to correct the residual error on curved surfaces. Finally, on the curved surfaces, we observed that the distribution of support forces could change quite significantly for some shapes as they changed orientation. For example, we found that the blue square prism could change from being supported stably on two opposite edges, to pivoting/rocking on two diagonally opposite corners as it was pushed across the concave surface. This effect can make it much harder to stabilize the motion of the object during pushing.

Another common trend seen for all surfaces is that performance is generally better for the smaller, narrower geometric objects, particularly the yellow triangular prism and red square prism. This behaviour is consistent with the theoretical results derived in [38], which show that there is a broader region of stability on the pushed edge as the distance between the CoF and the pushed edge decreases relative to the radius of the curved pusher (and hence the object is easier to control). This might also account for the higher accuracy obtained with the triangular prism, where the CoF is generally much closer to the pushed edge.

A final observation on Experiment 2 is that oscillatory behaviour or transient instability is more likely to occur immediately after episodes where the pusher is in contact with an external corner of an object and pushing through the CoF towards the target. When this happens (and with a large-radius, compliant sensor like the TacTip it can happen for reasonably long periods) the corrective action produced by the controller will be close to zero. However, a minimal shift of the sensor to one side or the other can cause a significant rotation around the CoF and a correspondingly large corrective action or overshoot. This destabilizing effect appears to get more pronounced closer to the target.

We also demonstrated that the method works for pushing irregular everyday objects across a flat horizontal surface. Despite differences in object shape, material and initial orientation (with respect to the pusher/sensor), the mean $\pm$ standard deviation of $y_{targ}$ was found to be $3.05 \pm 3.36$ mm across the five objects and ten random initial orientations. For some objects, such as the spray bottle, the path taken to the target shows a greater degree of variation than the paths taken by other objects, but the system still manages to find a stable pushing path towards the target.

In the literature, object pushing has been studied in a wide variety of contexts and with different types of robot. A wheeled mobile robot with an onboard camera was used in [8], and a Willow Garage PR2 robot with a Microsoft Kinect RGB-D camera attached to its head was used in [7]. Various different push movements have also been considered (e.g., overhead, fingertip or side-sweep in [7]) with objects that are matched to the experimental conditions. In [7], for example, many of the objects are low profile, while in [8] the objects tend to be larger and better-suited to pushing by a wheeled robot. Also, previous work has tended to use computer vision techniques to *estimate global properties* such as the centroid and (in some cases) the orientation of the object to control the push direction using perpendicular movements. In contrast, here we have used tactile sensing to *explore local properties* of the object, i.e. feel around objects, to find a stable pushing path. Nevertheless, to provide some measure of comparative accuracy of our method, we note that in [7] pushes were considered successful if they approached the target to within 20 mm or 40 mm (two levels of accuracy were used), and in [8] they were considered successful if they approached the target to within 50 mm.

We think there are three main limitations to our current method: the need to know the sensor contact angle in the vertical plane to keep the sensor moving parallel to the surface; the need for a target approach zone to avoid instability in that region; and the need to trade off controller goals in some situations. In future work, we aim to address these issues as follows. First, by using the shear detection properties of the sensor, it may be possible to detect the degree to which pushes are parallel to the surface and then use this information to regulate this aspect of behaviour. Second, by investigating more sophisticated gain-scheduling methods, we should be able to dispense with the target approach zone. Third, by relaxing the requirement for the sensor to remain perpendicular to the surface at all times and only requiring it to stay within the friction cone, this may help to reduce problems with incompatible controller goals. Finally, the current design of our tactile sensor and its attachment to the robot arm means that we cannot push low-profile, flat objects, which we also aim to address in future work.

In this work, we have demonstrated that it is possible to perform well at robotic pushing without directly modelling the dynamics of pusher-object interaction. Although we have used a model in our system (the CNN pose prediction model), we have not used a model that explicitly tries to predict how an object pose changes in response to a specific push action. This raises the interesting question of what aspects of the system should be modelled (regardless of the modelling approach) to achieve the best performance. To date, most of the effort has focused on modelling the dynamics of object interaction, but it might turn out that this is not the easiest aspect to generalize to different situations. We have also shown that while pushing over curved surfaces is more challenging than flat horizontal surfaces, it is still possible to achieve a reasonable level of performance using a simple approach. Since most real-world applications are unlikely to be restricted to flat horizontal surfaces, we think it is important that future research in this area also considers non-planar surfaces and other more complex environments when developing new methods.

### Acknowledgement

The authors thank Andy Stinchcombe and other members of the BRL Tactile Robotics group. This work was supported by an award from the Leverhulme Trust on 'A biomimetic forebrain for robot touch' (RL-2016-39).